\RequirePackage{tikz}
\usetikzlibrary{arrows}
\documentclass[pdflatex]{sn-jnl}

\usepackage{amssymb}
\usepackage{amsthm}
\usepackage{multicol}
\usepackage{multirow}
\usepackage{booktabs}
\usepackage{mathtools}
\usepackage{bbm}
\usepackage{tablefootnote}
\usepackage{caption} 
\usepackage[round]{natbib}
\newtheorem{theorem}{Theorem}[section]
\newtheorem{lemma}{Lemma}[section]
\theoremstyle{remark}
\newtheorem{remark}{Remark}[section]
\theoremstyle{definition}
\newtheorem{definition}{Definition}[section]

\DeclarePairedDelimiterX{\norm}[1]{\lVert}{\rVert}{#1}
\DeclareMathOperator*{\argmin}{arg\,min}

\providecommand{\keywords}[1]{\textbf{\textit{Keywords---}} #1}

\title[Improving state estimation through projection post-processing]{Improving state estimation through projection post-processing for activity recognition with application to football}

\author*[1]{Micha\l{} Ciszewski}
\author[1]{Jakob S\"{o}hl}
\author[1]{Geurt Jongbloed}
\affil[1]{\textit{Applied Mathematics, Delft University of Technology}, \textit{Mekelweg 4}, \textit{Delft}, \textit{2628 CD}, \textit{The Netherlands}}
\affil*{Corresponding author, M.G.Ciszewski@tudelft.nl}
\abstract{The past decade has seen an increased interest in human activity recognition based on sensor data. Most often, the sensor data come unannotated, creating the need for fast labelling methods. For assessing the quality of the labelling, an appropriate performance measure has to be chosen. Our main contribution is a novel post-processing method for activity recognition. It improves the accuracy of the classification methods by correcting for unrealistic short activities in the estimate. We also propose a new performance measure, the Locally Time-Shifted Measure (LTS measure), which addresses uncertainty in the times of state changes. The effectiveness of the post-processing method is evaluated, using the novel LTS measure, on the basis of a simulated dataset and a real application on sensor data from football. The simulation study is also used to discuss the choice of the parameters of the post-processing method and the LTS measure.}


\date{}

\begin{document}

\maketitle


\section{Introduction}


In almost all areas of science and technology, sensors are becoming more prevalent. In recent years we have seen applications of sensor technology in fields as diverse as energy saving in smart home environments \citep{lima15}, performance assessment in archery \citep{eckelt20}, detection of mooring ships \citep{waterbolk19}, early detection of Alzheimer disease \citep{varatharajan18} and recognition of emotional states \citep{kolakowska20}, to name just a few.


Our main interest lies in the detection of human activities using sensors attached to the body.
Sensors generate unannotated raw data, suggesting the use of unsupervised learning methods.
If an activity specified in advance is of interest, then supervised learning and labelled data are required.
However, the task of labelling activities manually from sensor data is labour-intensive and prone to errors, which creates the need for fast and accurate automated methods.


Human activity recognition (HAR) attracted much attention since its inception in the~'90s.
A plethora of methods are currently being used to detect human activities \citep{lara13}, with various deep learning techniques leading the charge \citep{minhdang20, wang19}.
In many studies \citep{ronao17,capela15,avilescruz19} only sensors embedded in a smartphone are used to classify user activities.
Physical sensors, such as accelerometers or gyroscopes attached directly to a body or video recordings (from a camera), are the most popular sources of data for activity recognition \citep{rednic12,zhu11,cornacchia16}.
Similarly, cameras can be either placed on the subject \citep{li11,ryoo13,watanabe11} or they can observe the subject \citep{song11,laptev08,ke05}.
Rarely, both camera and inertial sensor data are captured at the same time \citep{chen15}.


The temporal structure of the time series should be taken into account when choosing a method for activity recognition.
Simple classification techniques (such as logistic regression or decision trees) ignore time dependencies and will need to be improved after the procedure.
Alternatively, methods which are more complicated and more difficult to train have to be deployed.
Another challenge lies in the reliability of manual labelling (in case of supervised learning).
Quite often it is unreasonable to assume that labels annotating the observed data are exact with regards to timings of transitions from one activity to another \citep{ward06}.
Timing uncertainty can be caused by a deficiency of the manual labelling or the inability to objectively detect boundaries between different activities.
This issue is well-known in the literature, for instance, \citet{yeh17} introduced a scalable, parameter-free and domain-agnostic algorithm that deals with this problem in the case of one-dimensional time series.


The main contribution of this paper is the introduction of a post-processing procedure, which improves a result of activity classification by eliminating too short activities.
The method requires a single parameter which can be interpreted as the minimum duration of the activites (hence the choice of this parameter is driven by domain knowledge).
It allows us to mitigate the problem of activites being fragmented in cases where some domain-specific information about state durations is available.
Based on empirical evidence, the performance of classical machine learning classifiers improves significantly by our post-processing method. 
This enables simple and fast but less accurate classification methods to be upgraded to accurate and fast classifiers.


In order to compare the quality of competing activity recognition methods, an appropriate criterion for evaluating the performance is needed (also to demonstrate the performance of the post-processing procedure we introduce).
Below are some commonly used performance measures:
\begin{itemize}
\item accuracy, precision, the $F$-measure \citep{lara13, lima19},
\item similarity measures for time series classification \citep{serra14}, such as Dynamic Time Warping or Minimum Jump Costs Dissimilarity,
\item custom vector-valued performance metric \citep{ward11}.
\end{itemize}
Our objective is to design a performance measure that satisfies problem-specific conditions, which will be specified later.


The outline of the paper is as follows. 
Section 2 provides a method for improving classification with a post-processing scheme that uses background knowledge on the specific context.
In particular, it validates the state durations and provides an improved classification that satisfies the physical constraints on the state durations imposed by the context. 
Section 3 introduces specialized performance measures for assessing the quality of classification in general and in activity recognition in particular.
The new performance measure is also designed to show the advantages of the post-processing fairly.
Section 4 presents an application of the techniques in a simulated setting. The post-processing method was able to improve the estimates significantly. The method achieves similar results in an application to football data.

\section{Improving classification by imposing physical restrictions}\label{PPsect}

\subsection{Post-processing by projection}\label{sect:ppintro}

When recognizing human activities, it is often the case that the result of the classification contains \emph{events} (time intervals in which a classification result is constant) that are too short\footnote{Depending on the application `too short' might be specified differently.}.
Usually ad hoc methods are used in order to discard those events, e.g. removal of any short events and replacing them with the next state in the classification, whose length is above a fixed threshold.
The goal of this section is to introduce a formalized approach to correcting for the classifier's mistakes regarding the distribution of durations by introducing a novel post-processing procedure.

Consider the set of states $\mathcal{S}=\{1,...,M\}$ and a metric $d$ on $\mathcal{S}$.
Let $\rho$ denote the \emph{discrete metric}\footnote{Distance between two different states is equal 1 and distance from a state to itself is equal 0.} on $\mathcal{S}$.
Any states-valued function of time will be called a \emph{state sequence}.
In reality we are only able to obtain a discrete-time signal, however, the relevant information contained in such a signal is a list of all the state transitions, which can more easily be encoded in a function with continuous argument.
Hence, we define~$\mathcal{T}$, the set of all càdlàg\footnote{right continuous, left limits exists} functions $f:\mathbb{R}\rightarrow\mathcal{S}$ with a finite number of discontinuities.
We define the \emph{standard distance} induced by a metric $d$ between two state sequences as
\begin{equation}\label{eq:standarddistanceonmathcalt}
\textrm{dist}:\mathcal{T}\times\mathcal{T}\ni (f,g)\rightarrow \textrm{dist}(f,g)=\int\limits_\mathbb{R}d(f(t),g(t))dt.
\end{equation}
If $d$ is a metric on $\mathcal{S}$, then $\textrm{dist}$ is a metric on $\mathcal{T}$.
The standard distance induced by the discrete metric is the time spent by $f$ in a state different from $g$.

Now, we define a measure of closeness between functions in $\mathcal{T}$, as our goal is to find a function close enough to a given function in $\mathcal{T}$, while reducing the number of jumps it has (which in turn will eliminate short events in the state sequence).
Let $f,g\in\mathcal{T}$. Then we introduce the notation:
\begin{equation}
\label{approxscheme}
E_\gamma(f,g)=\textrm{dist}(f,g) + \gamma\cdot \vert J(g)\vert,
\end{equation}
where $J(g)$ is the set of all discontinuities of $g$, $\vert J(g)\vert$ is the number of all discontinuities of $g$ and $\gamma$ is a penalty for a single jump of $g$.

Given $f\in\mathcal{T}$, our goal is to find any solution $\hat{f}\in\mathcal{T}$ of the minimization problem
\begin{equation}\label{eq:projectiondist}
\hat{f}\in\argmin\limits_{g\in\mathcal{T}}E_\gamma(f,g).
\end{equation}
As a default, we will use the standard distance induced by the discrete metric.

In order to characterize the solution $\hat{f}$ of problem \eqref{eq:projectiondist} we present the following lemma.

\begin{lemma}\label{jumpsoutside}
Let $\gamma>0$ and $f\in\mathcal{T}$. Let $J$ denote the set of all discontinuities of the function $f$. 
There exists a solution $\hat{f}$ of the problem \eqref{eq:projectiondist} such that it does not contain jumps outside of $J$.
\end{lemma}

Lemma \ref{jumpsoutside} leads to the conclusion that in search for the solution of the minimization problem we can limit ourselves to a finite set of functions, namely a subset of $\mathcal{T}$ with jumps only allowed at the same locations as function $f$. 
The proof of lemma \ref{jumpsoutside} can be found in the appendix.

In this minimization problem the choice of the parameter $\gamma$ plays a crucial role.
We will now show an interpretation of the penalty parameter that will ease the process of choosing it. 
It will also allow us to reformulate problem \eqref{eq:projectiondist}. 
First, we define a new set of functions.
\begin{definition}[Function with bounded minimum duration of states]
Given a parameter $\gamma>0$ we define $\mathcal{G}_\gamma\subset\mathcal{T}$, the set of functions with \textit{bounded minimum duration of states}, such that for $g\in\mathcal{G}_\gamma$ we have
\begin{itemize}
\item $g=\sum\limits_{i=1}^{n-1} s_i\mathbbm{1}_{[t_i,t_{i+1})}$ for some constant $n\in\mathbb{N}$, a sequence of states $\{s_1,...,s_{n-1}\}$, such that $s_i\neq s_{i+1}$ for $i=1,...,n-2$, and an increasing sequence $t_1< t_2< ...< t_n$ (we allow $t_1=-\infty$ and $t_n=\infty$),
\item if $n\geq 2$, then $\forall_{i\geq 2}\quad t_i-t_{i-1} \geq \gamma$.
\end{itemize}
\end{definition}

Lemma \ref{gammainterpret} yields a connection between the penalty $\gamma$ and the minimum duration of states that we impose on the solution of our minimization problem.
\begin{lemma}\label{gammainterpret}
Let $\gamma>0$ and $f\in\mathcal{T}$. Any solution $\hat{f}$ of problem \eqref{eq:projectiondist} is an element of $\mathcal{G}_\gamma$.
\end{lemma}

This lemma can be used in practice to select the size of the penalty.
The proof of lemma \ref{gammainterpret} can be found in the appendix.

Given $f\in\mathcal{T}$, the minimization problem \eqref{eq:projectiondist} is equivalent to the minimization problem
\begin{equation}\label{eq:projectionproblem}
\hat{f}\in\argmin\limits_{g\in\mathcal{G}_\gamma}E_\gamma(f,g)
\end{equation}
by lemma \ref{gammainterpret}.
$\hat{f}$ will be called a projection of $f$ onto $\mathcal{G}_\gamma$.

As mentioned before, the regularization by penalizing high numbers of jumps narrows down the set of possible solutions to a finite nonempty subset of $\mathcal{G}_\gamma$ (thanks to lemma \ref{jumpsoutside}), which leads to the existence of~$\hat{f}$. 
However, the solution might not be unique, as illustrated by the following example. 

Consider $\mathcal{S}=\{0,1\}$, $f=\mathbbm{1}_{[0.35,0.45)}+\mathbbm{1}_{[0.55,+\infty)}$ and $\gamma=0.2$. 
Both~$\hat{f}_1=\mathbbm{1}_{[0.35,+\infty)}$ as well as $\hat{f}_2=\mathbbm{1}_{[0.55,+\infty)}$ are the projections of $f$. 
One could think of it as an issue, however, it does reflect well our understanding of the original problem.
The assumption is that~$f$ has impossibly short windows, because it is uncertain which activity is actually performed in the interval $[0.35, 0.55)$.
Looking only at~$f$ we are unable to decide which solution is more suitable, hence it is only natural that the method also returns two possible options.

We close with a remark regarding influence of the extreme values of $\gamma$ on projection $\hat{f}$.
\begin{remark}
Let $f\in\mathcal{T}$.
If $\gamma=0$, then $\hat{f}=f$ is the only projection of $f$.
If $\gamma=\infty$ and $E_\gamma(f,g)<\infty$ for some function $g\in\mathcal{T}$\footnote{Note that this is not always true. If the first and the last states of $f$ are different, then any function can be a projection of $f$.}, then $g$ is constant and equal everywhere to the most common state of $f$ and $\hat{f}=f$\footnote{Note that if $E_\gamma(f,g)<\infty$, then the first and the last states of $f$ are the same and the constant function equal to that state is the only projection}.
\end{remark}

\subsection{Connection with the shortest path problem}

In this section we devise a method for finding a projection in an efficient manner.
It will be shown that the problem of finding the shortest path in a particular graph is equivalent to the minimization problem \eqref{eq:projectionproblem}.
This is possible thanks to the lemmas \ref{jumpsoutside} and \ref{gammainterpret}, which narrowed down the set of possible solutions to a finite set.

First, we present a lemma which further characterizes a projection of $f$.
\begin{lemma}\label{longstates}
Let $f\in\mathcal{T}$. 
Suppose $f\equiv c$ on an interval $[a,b]$ for some constant $c\in\mathbb{R}$.
If $b-a>2\gamma$, then $\hat{f}\equiv c$ on $[a,b]$.
If $b-a=2\gamma$, then there exists a projection such that $\hat{f}\equiv c$ on $[a,b]$.
\end{lemma}
The proof of lemma \ref{longstates} can be found in the appendix.
\begin{remark}\label{simplefact}
If $n>2$ and all states are shorter than $2\gamma$ (with exception of the first and the last state), there exists a projection such that the second and the second-to-last jump of the original function are not present in it.
\end{remark}
Remark \ref{simplefact} allows us to ignore the second and the penultimate jump of the original function when searching for jump locations in the projection. The proof of this remark can be found in the appendix.

Without loss of generality we will assume that~$f$ has $n\geq 2$ jumps at time points $t_i$ for $i=1,...,n$:
\begin{equation*}
f=\sum\limits_{i=0}^n s_i\mathbbm{1}_{[t_i,t_{i+1})},
\end{equation*}
where $s_i\in\mathcal{S}$ for $i=0,...,n$ and $s_i\neq s_{i+1}$ for $i=0,...,n-1$. We use the following notation: $t_0=-\infty$, $t_{n+1}=\infty$. 
In light of lemma \ref{longstates} we assume that
\begin{equation}\label{eq:split}
t_{i+1}-t_i<2\gamma
\end{equation}
for $i=1,...,n-1$.
If this is not the case, then the function needs to be split up into several parts in such a way that in each part equation \eqref{eq:split} holds.

We will now define a graph for the purpose of showing the connection between the problem of finding a projection $\hat{f}$ and the problem of finding a shortest path in a directed graph.
Let $G=(V,A)$ be a directed graph such that the set of vertices $V$ is given by
\begin{equation}\label{eq:vertices}
V=\{t_0,t_1,...,t_n,t_{n+1}\}\backslash\{t_2,t_{n-1}\}\footnote{In case of $n=2$, both jumps have to be included in the set of vertices.}
\end{equation}
and the set of directed arcs is given by
\begin{equation}\label{eq:arcs}
A=\{(t_k,t_l)\in V^2:t_l-t_k\geq \gamma\}.
\end{equation}
There is a correspondence between each path from $t_0$ to $t_{n+1}$ and a sequence of jumps in the interval $(t_1-\gamma,t_n+\gamma)$. 
A path $(t_0,t_{l_1},...t_{l_m},t_{n+1})$ can be associated with a function $g$ with jumps at $t_{l_1}$, ..., $t_{l_m}$, such that $g(t_{l_k})$ is the most common value of $f$ in interval $[t_{l_k},t_{l_{k+1}})$. 
The definition \eqref{eq:arcs} of the set of directed arcs ensures that all paths in the graph $G$ correspond to at least one function in $\mathcal{G}_\gamma$.

We now introduce a weight function $W:A\rightarrow\mathbb{R}_+$ ensuring that the cost of the path coincides with the error $E(f,\cdot)$ of the corresponding function in the interval $(t_1-\gamma,t_n+\gamma)$.
Let $I_k=t_{k+1}-t_k$ for $k=0,...,n$. 
It is noteworthy that $I_0, I_n=\infty$, while $I_k<2\gamma$ for $k=1,...,n-1$.
We introduce the penalty for a jump~$\phi_k=\gamma$ for $k=1,...,n$ and $\phi_{n+1}=0$.
Now we define the weight function~$W$:
\begin{equation}\label{eq:weight}
W((t_k,t_l))=\sum\limits_{m=k}^{l-1}I_md(s_{kl},s_m)+ \phi_l,
\end{equation}
for $(t_k,t_l)\in A$, where $s_{kl}$ represents the most common state in the interval $[t_k,t_l)$ of the original function $f$.
The first term equals the~$\mathrm{dist}(f,g)$ in $[t_k,t_l]$.
The second term adds a penalty for jump at $t_l$ if $t_l$ is finite (the penalty for jump at $t_k$ was added on a previous arc in the path, if $k>0$).

\begin{theorem}[Problem equivalence]\label{equivtheorem}
Let $\gamma>0$ and $(t_1,...,t_n)$ be the only discontinuities of a function $f\in\mathcal{T}$. 
Let $G=(V,A,W)$ be a weighted, directed graph as defined in \eqref{eq:vertices}, \eqref{eq:arcs}, \eqref{eq:weight} above.
The task of finding a projection of~$f$ onto~$\mathcal{G}_\gamma$, as defined in \eqref{eq:projectionproblem}, is equivalent to finding the shortest path from~$t_0$ to~$t_{n+1}$ in the graph~$G$.
\end{theorem}

The proof of the theorem can be found in the appendix.
Now, we will illustrate the method by an example.

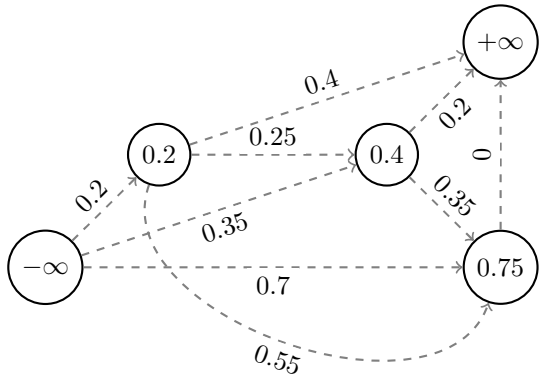
\begin{figure}[htbp!]
    \centering
    \begin{tikzpicture}
    \begin{scope}[every node/.style={circle,thick,draw}]
        \node (A) at (-3,0) {$-\infty$};
        \node (B) at (-1.5,1.5) {$0.2$};
        \node (C) at (1.5,1.5) {$0.4$};
        \node (D) at (3,0) {$0.75$};
        \node (E) at (3,3) {$+\infty$};
    \end{scope}
    \begin{scope}[every node/.style={sloped, fill=white, auto=false},
                  every edge/.style={draw=gray,dashed,thick}]
        \path [->] (A) edge node [above] {$0.2$} (B);
        \path [->] (A) edge node [below] {$0.35$} (C);
        \path [->] (A) edge node [below] {$0.7$} (D);
        \path [->] (B) edge node [above] {$0.25$} (C);
        \path [->] (B) edge [bend right = 90] node [below] {$0.55$} (D);
        \path [->] (B) edge node [above] {$0.4$} (E);
        \path [->] (C) edge node [above] {$0.35$} (D);
        \path [->] (C) edge node [below] {$0.2$} (E);
        \path [->] (D) edge node [above] {$0$} (E);
    \end{scope}
    \end{tikzpicture}
    \caption{Graph $G$ constructed for the function $f$}
    \label{fig:graph}
\end{figure}

Given $\gamma=0.2$ and $\mathcal{S}=\{0,1,2,3\}$, consider the function $f=\mathbbm{1}_{[0.2, 0.35)}+2\cdot\mathbbm{1}_{[0.4, 0.55)}+3\cdot\mathbbm{1}_{[0.55,0.75)}+2\cdot\mathbbm{1}_{[0.75,+\infty)}$.
The graph $G$, as defined in \eqref{eq:vertices}, \eqref{eq:arcs}, \eqref{eq:weight}, for~$f$, is shown in figure \ref{fig:graph}.

There are seven possible paths from $-\infty$ to $+\infty$.
The path $\hat{P}=(-\infty,0.4,\infty)$ has the cost equal to $0.55$ and is the shortest path from $-\infty$ to~$+\infty$.
Hence we conclude that $\hat{f}=2\cdot\mathbbm{1}_{[0.4, \infty)}$ is the projection of $f$ onto $\mathcal{G}_{0.2}$ (in this case, it can be shown $\hat{f}$ is the only projection of $f$).

\subsection{Binary case}

In case the set of states $\mathcal{S}$ consists of only two elements, a stronger result than lemma \ref{gammainterpret} can be achieved.
The main advantage of the binary case comes from the fact that we do not need to specify the sequence of states since knowing the starting state, each jump signifies a move to the only other available state.
First, we present a supporting remark which further strengthens the relation between jumps of a function from $\mathcal{T}$ and its projection.

For the remainder of the section, we will always assume that $\mathcal{S}=\{0,1\}$\footnote{This convention conflicts the notation established in section \ref{sect:ppintro} as it is more natural to use 0 and 1 as states in the binary case.}.
\begin{lemma}\label{binarydirections}
Let $\gamma>0$ and $f\in\mathcal{T}$. Let $J$ denote the set of all discontinuities of the function $f$. 
If a function $g\in\mathcal{G}_\gamma$ contains a jump $j\in J(f)$, but in an opposite direction than in $f$, then $g$ cannot be a projection of $f$ onto $\mathcal{G}_\gamma$.
\end{lemma}

\begin{lemma}\label{binarygammainterpret}
Let $\gamma>0$ and $f\in\mathcal{T}$. Any solution $\hat{f}$ of the problem \eqref{eq:projectiondist} is an element of $\mathcal{G}_{2\gamma}$.
\end{lemma}

The proofs of remark \ref{binarydirections} and lemma \ref{binarygammainterpret} can be found in the appendix.
Lemma \ref{binarygammainterpret} leads to the equivalence of the problem \eqref{eq:projectionproblem} with the minimization problem:
\begin{equation}\label{eq:binaryprojectionproblem}
\hat{f}\in\argmin\limits_{g\in\mathcal{G}_{2\gamma}}E_\gamma(f,g).
\end{equation}
The strengthening of lemma \ref{jumpsoutside} by restricting not only the locations of the jumps but also their directions is a favorable change as it narrows the set of possible solutions.

\begin{lemma}\label{binarylongstates}
Let $f\in\mathcal{T}$. 
Suppose $f\equiv c$ on an interval $[a,b]$ for some constant $c\in\mathbb{R}$.
If $b-a>\gamma$, then $\hat{f}\equiv c$ on $[a,b]$.
If $b-a=\gamma$, then there exists a projection such that $\hat{f}\equiv c$ on $[a,b]$.
\end{lemma}

Lemma \ref{binarylongstates} potentially reduces the number of jumps that have to be considered in the post-processing. 
Moreover, remark \ref{binarydirections} reduces the number of arcs when building the graph making the process of finding the shortest path more effective.

The directed graph $G$ has a different set of directed arcs compared to \eqref{eq:arcs}:
\begin{equation}\label{eq:arcsbinary}
A=\{(t_k,t_l)\in V^2:t_l-t_k\geq2\gamma\; \textrm{and}\; l-k\bmod 2\equiv 1\}.
\end{equation}

\begin{theorem}[Problem equivalence - binary version]\label{binaryequivtheorem}
Let $\gamma>0$ and $(t_1,...,t_n)$ be the only discontinuities of a function $f\in\mathcal{T}$. 
Let $G=(V,A,W)$ be a weighted, directed graph as defined in \eqref{eq:vertices}, \eqref{eq:arcsbinary}, \eqref{eq:weight}.
The task of finding a projection of~$f$ onto~$\mathcal{G}_{2\gamma}$, as defined in \eqref{eq:binaryprojectionproblem}, is equivalent to finding a shortest path from~$t_0$ to~$t_{n+1}$ in the graph~$G$.
\end{theorem}

Proofs of lemma \ref{binarylongstates} as well as the theorem \ref{binaryequivtheorem} can be found in the appendix.

\section{Incorporating domain knowledge into the performance measure of classification}

\subsection{Problem-specific requirements on the performance measure}\label{firstrealsect}

In order to choose an appropriate performance measure for a given classification task, it is important to understand the problem-specific demands on the result.
The standard distance \eqref{eq:standarddistanceonmathcalt}, which can be understood as a continuous analogue of the most common performance metric, namely the misclassification rate, can often be inadequate to compare the classification results as it is a one-fits-all type of metric and if more is known about the problem, it might not represent the idea of accuracy that users have in mind.
On the other hand, there have been other approaches to performance metrics, e.g. \citep{ward11}.
Their approach focuses on characterizing the error in terms of the number of inserted, deleted, merged and fragmented events.
Event fragmentation occurs when an event in the true labels\footnote{If a state sequence corresponds to the true underlying sequence of activities in a time series, then it will be called the \emph{true labels}} is represented by more than one event in the estimated labels\footnote{An estimate of the true labels will be called the \emph{estimated labels}.}, whereas merging refers to several events in true labels being represented by a single event in the estimated labels.
\citet{ward11} provide an overview of different performance metrics used in activity recognition proposing a solution to the problem of timing uncertainty as well as event fragmentation and merging.
Their solution is based on segments, which are intervals in which neither the true labels nor the estimated labels change the state.
If the state in the estimate and the state in the true labels agree in a given segment, they denote it as correctly classified.
If that is not true, the segment is classified accordingly as fragmenting segment, inserted segment, deleted segment or merged segment.
This provides a deeper level of error characterization, which is then used in different metrics of classifier performance.
Their vector-valued performance metric is preferable when in-depth overview of the types of mistakes made by the classifier is needed.
We will introduce a novel scalar-valued performance metric, which can be easily compared and includes problem-specific information such as timing uncertainty in the labels.

In this section, we aim at highlighting the main characteristics of the classification of movements based on wearable sensors and at translating them into specific requirements on the performance measure.
Our first requirement comes from physical restrictions.
The states considered in our application represent human activities, but also in more general contexts they often cannot be arbitrarily short; there is a lower bound on the length of the events in a state sequence.
Hence, estimated labels that violate this lower bound indicate a bad performance.
The lower bound condition requires two parameters: the lower bound and the penalty for each violation.
The lower bound can either be estimated or determined from domain knowledge, while the penalty can be chosen more freely.
Through physical restrictions we can see a deeper connection with the method introduced in section \ref{PPsect}.
It is clear that the standard classification methods cannot ensure that the state sequence contains only events longer than a certain level.
The post-processing method addresses this issue directly and as a consequence we can expect classifiers to benefit from it in the context of the new performance measure.

The issue of timing uncertainty should also be addressed when designing the performance measure.
To illustrate its importance more clearly, we present an example.
Five people were asked to detect boundaries between activities in different time series using a visualization tool.
The tool outputs an animated stick figure model\footnote{A symbolic representation of the human body using only lines} given sensor data.

Three time series were selected, each with one of the following activities: running, jumping and ball kick.
The start and the end of each activity were recorded by participants.
Table \ref{tab:stickmodel} presents the results of the experiment.

\begin{table}[htbp!]
\begin{center}
\begin{tabular}{lcccccc}
\toprule
\multirow{2}{*}{Partic.} & \multicolumn{2}{c}{Running} & \multicolumn{2}{c}{Jumping} & \multicolumn{2}{c}{Kick} \\
& Start & End & Start & End & Start & End \\
\midrule
P1 & 2 & 7.3 & 2.7 & 5.2 & 2.5 & 3.5 \\
P2 & 2 & 7.5 & 2.7 & 5.2 & 2.5 & 3.9 \\
P3 & 2.3 & 6.6 & 2.7 & 5.1 & 2.7 & 3.6 \\
P4 & 2.3 & 7.2 & 2.7 & 5.3 & 2.5 & 4.3 \\
P5 & 2.2 & 7.2 & 2.9 & 5.4 & 2.5 & 4.1 \\
\midrule
Avg. & 2.16 & 7.16 & 2.74 & 5.24 & 2.54 & 3.88 \\
Std & 0.15 & 0.34 & 0.09 & 0.11 & 0.09 & 0.33 \\
\bottomrule
\end{tabular}
\end{center}
\caption{The results of the labelling experiment; all times are in seconds. The two last rows show the average and the sample standard deviation for each boundary}
\label{tab:stickmodel}
\end{table}

The experiment indicates there is indeed uncertainty regarding the state transitions.
Granted that the sample size is very small, we notice more variation in results referring to the end of activities rather than the beginnings.
Additionally, we see more variation in the results for the kick than the jumping.
So the boundaries of some activities seem to be more difficult to identify than of others.

\subsection{Globally Time-Shifted distance}\label{gtssect}

The standard distance \eqref{eq:standarddistanceonmathcalt} is an unsatisfying measure to compare two state sequences, since it does not incorporate the requirements posed in the previous section.
In order to improve it, we start by modelling the timing uncertainty.
Let $f\in\mathcal{T}$ be the true labels process and let $f$ have $n$ discontinuities $t_1,...,t_n$.
The locations of the discontinuities are corrupted by additive noise:
\begin{equation*}
    t_i=T_i+X_i,
\end{equation*}
for all $i=1,...,n$, where $T_i$ is the true and unknown location of the $i$-th jump.
In this section we will assume that $X_1=X_2=...=X_n$ (all jumps are moved by the same value; the global time shift), although in general, it is more realistic to assume that $X_1,...,X_n$ are independent random variables.
We will relax this condition later.

We define a class of Globally Time-Shifted distances (GTS distances), loosely inspired by the Skorokhod distance on the space of càdlàg functions \cite[pp.~121]{billingsley99}.
The GTS distances are parametrized by two parameters. 
The parameter $w$ controls the weight of misclassification occurring from the uncertainty of the true labels, while the parameter~$\sigma$ controls the magnitude of the shift of activities.
\begin{definition}[Globally Time-Shifted distance]
Let $f,g\in\mathcal{T}$.
Given $w\geq 0,\sigma>0$ and a metric $d$ on $\mathcal{S}$ we define a Globally Time-Shifted distance as:
\begin{equation*}
GTS_{w,\sigma}(f,g)=\inf\limits_{\epsilon\in[-\sigma,\sigma]}\{\textrm{dist}(f\circ\tau_\epsilon,g)+w\vert\epsilon\vert\},
\end{equation*}
where for $\epsilon>0$ $\tau_\epsilon:\mathbb{R}\rightarrow\mathbb{R}$ is a time shift defined as follows:
\begin{equation*}
\tau_\epsilon(t)=t-\epsilon.
\end{equation*}
\end{definition}

Depending on the choice of parameters the GTS distance possesses certain properties.
For $w>0$ and $\sigma=\infty$, the GTS distance is an extended metric\footnote{It may attain the value $\infty$.} and a proof of this fact is given in the appendix.
If $w>0$ and~$\sigma>0$, then it is a semimetric meaning that it has all properties required for a metric, except for the triangle inequality.

The main downside of using the GTS distance is the unrealistic assumption on timing uncertainty.
However, if we know that the true labels preserve the true state durations then it is a good choice.
Consider a function~$f\in\mathcal{T}$ with two state transitions located at $t_1$ and $t_2$.
Let $g\in\mathcal{T}$ also feature two state transitions located at $t_1 - \tau_1$ and $t_2 - \tau_2$.
If $\tau_1\neq\tau_2$, then there is no global time shift that can align the functions $f$ and~$g$.
This implies that the true state durations need to be preserved in the estimate in order to align functions using the global time shift.

\subsection{Locally Time-Shifted distance and the Duration Penalty Term}

\begin{figure}
    \centering
    \includegraphics[width=\linewidth]{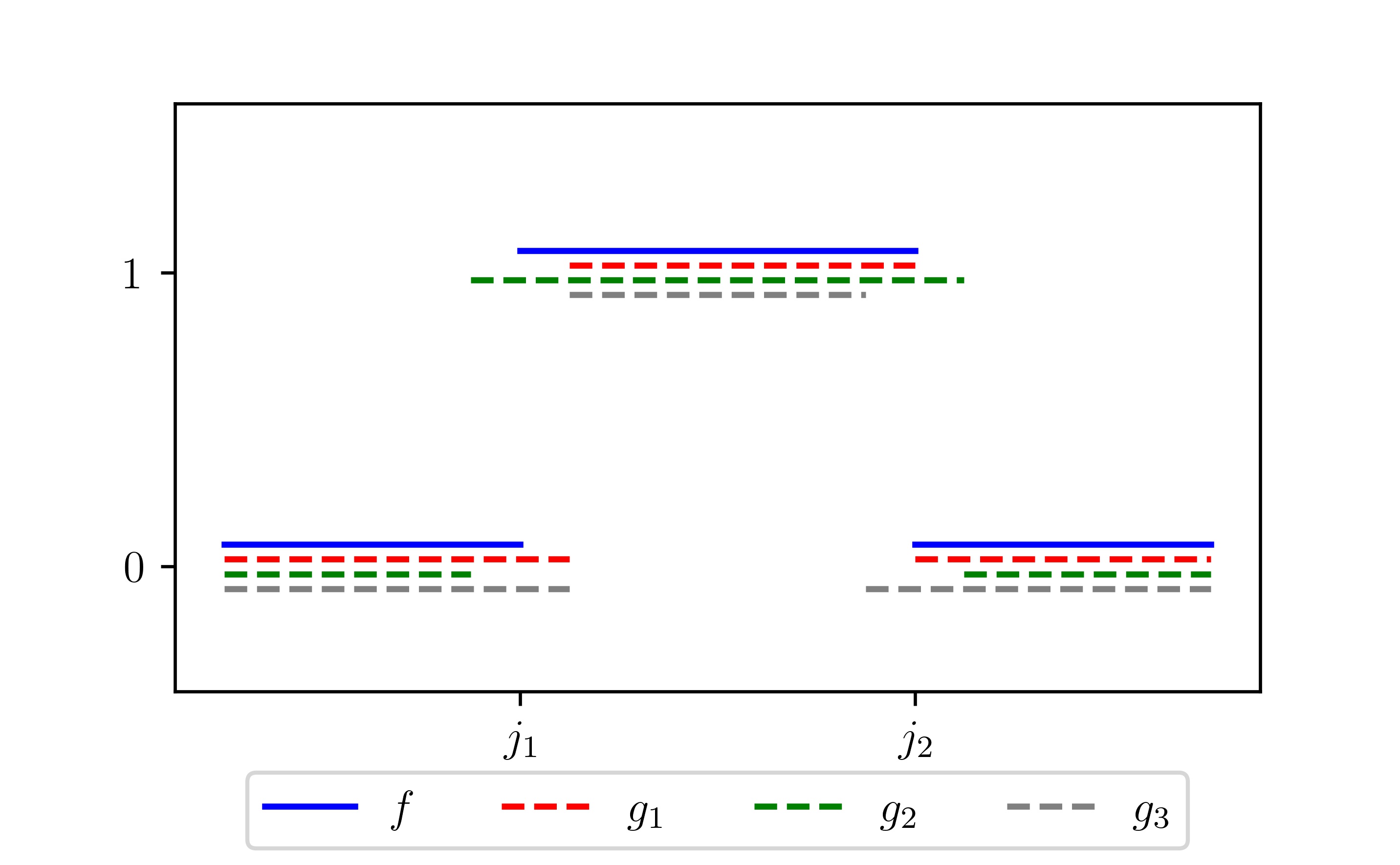}
    \caption{The function $f$ represents the true labels with an uncertainty around state boundaries, $g_i$ are the approximations of $f$}
    \label{fig:example231}
\end{figure}

The global time shift stresses the state durations, which is not always desirable.
For instance, if the true labels do not preserve the real state durations, or e.g. if the additive noise terms in the locations of the jumps are independent.
Here is an example: figure \ref{fig:example231} shows~$f$ and its approximations~$g_i$ for $i=1,2,3$.
It is impossible to align~$f$ with any of the~$g_i$ with a single time shift, however, it would be possible if each state transition could be shifted `locally'.

Naturally, to accommodate for this issue, a suitable modification would be to replace one global time shift with multiple local time shifts. 
We now introduce a measure of closeness between state sequeneces which conceptually can be seen as derived from the GTS measure.
We will be working with sequences of jumps, but more specifically given two sequences of state boundaries we will combine them together and sort the resulting joint sequence in an increasing order.
Subsequent pairs of values in this sequence are determining segments understood as in \citet{ward11}.
We weigh different types of segments and the result is a weighted average of segment lengths, which is supposed to reflect well the error magnitude of the classifier.

We define segments formally and introduce a new distance on $\mathcal{T}$.
\begin{definition}[Segments]\label{segments}
    Let $f, g\in\mathcal{T}$. 
    The elements of the smallest partition\footnote{A partition that cannot be made coarser} of $\mathbb{R}$ such that in each element of the partition neither $f$ nor $g$ changes state will be called segments.
\end{definition}
Since functions from $\mathcal{T}$ are piece-wise constant and have a finite number of discontinuities, there is always a finite number of segments.
The general form of segments that we will use is as follows:
\begin{equation}\label{eq:segments}
    (-\infty, a_1)\cup\bigcup\limits_{i=1}^{l-1}[a_i, a_{i+1})\cup[a_l,\infty),
\end{equation}
where $a_1< a_2...< a_l$ if $f$ and $g$ are not both constant on the real line.
Otherwise there is only one segment, consisting of the whole real line.
By convention, $a_0=-\infty$ and $a_{l+1}=\infty$, and
\begin{equation*}
    f(a_0)=f(a_1^{-})=\lim\limits_{x\rightarrow-\infty}f(x),\;f(a_{l+1})=f(a_l).
\end{equation*}
\begin{definition}[Locally Time-Shifted distance]
    Let $w\geq 0$, $\sigma>0$ and $d$ be a metric on $\mathcal{S}$. 
    Let $f,g\in\mathcal{T}$ and their set of segments to be denoted as in \eqref{eq:segments}. 
    We define the Locally Time-Shifted distance (LTS distance) as
    \begin{equation*}
        LTS_{w, \sigma}(f, g)=\sum\limits_{i=1}^{l-1}\delta_i(a_{i+1}-a_i)d(f(a_i),g(a_i)),
    \end{equation*}
    where
    \begin{equation*}
        \delta_i=
        \begin{cases}
            w, & a_{i+1}-a_i\leq \sigma,\; f(a_{i-1})=g(a_{i-1}), f(a_{i+1})=g(a_{i+1})\\
            1, & \text{otherwise}.
        \end{cases}
    \end{equation*}
\end{definition}
Similarly to the GTS distance, the parameter $w$ controls the weight of misclassification occurring from the uncertainty of the true labels.
The case when~$w < 1$ is more interesting to us, since it corresponds to timing uncertainty of the labels.
If $w\geq1$, then we put more importance on the timings of the jumps (opposite to timing uncertainty).
The LTS distance is an extended semimetric for $w>0$ (for a proof, see the appendix).
The triangle inequality does not hold in general.

The LTS distance addresses the issue of timing uncertainty in the true labels.
Let $\zeta>0$\footnote{Note that $\zeta$ is related in its interpretation to the $\gamma$ parameter introduced in section \ref{PPsect}.} be the lower bound on the lengths of the events as determined by the domain knowledge (or through estimation if possible).
Let~$\lambda>0$ be the penalty for each violation of the lower bound condition.
For~$f\in\mathcal{T}$ with its discontinuities $t_1,...,t_n$, we introduce a \emph{duration penalty term}:
\begin{equation*}
    DP_{\lambda,\zeta}(f) = \lambda\sum\limits_{k=1}^{n-1}\mathbbm{1}_{[0,\zeta)}(t_{k+1}-t_k).
\end{equation*}
This term will allow to lower the performance of classifications with unrealistically short events.

In practice, we will need to extend the functions to the real line in order to use the LTS distance as it is defined for functions with domain equal to the whole of~$\mathbb{R}$.
One natural extension could be to extend the first and the last state of each function indefinitely.
However, this solution leads to a problem.
Let $M>0$.
Consider two functions $f:[0,M]\rightarrow\mathcal{S}$ and $g:[0,M]\rightarrow\mathcal{S}$ such that for some $0<a<M$, $f(t)\neq g(t)$ on $[0,a)$.
No matter how small $a$ is, the distance between extended~$f$ and $g$ will always be infinite when using this extension, since in this case extended $f$ and $g$ are in different states on the whole half line $(-\infty,a)$.
Both functions need to be extended by the same state for the distance to be finite.
We extend any function~$f$ defined on interval $[0,M]$ to the real line, setting its value to an arbitrary state outside of $[0,M)$.
The distance is independent of the chosen state, as on the infinite segments that it introduces $f$ and $g$ are both equal.
Without loss of generality, we choose state $1$.
\begin{equation}\label{eq:extensionLTS}
f^\ast(t)=
\begin{cases}
f(t), & t\in[0,M)\\
1, & t\not\in[0,M).
\end{cases}
\end{equation}
Notice that this extension does not have the problem stated above as $f^\ast$ and $g^\ast$ are equal on the segments that it introduces and does not change the value on the original segments regardless of the choice of the state outside of $[0,M]$.

We combine the LTS distance and the duration penalty term to define the LTS measure of closeness of two state sequences.
\begin{definition}
Let $f$ be a function of true labels and $g$ its estimate, both defined on $[0,M]$.
The \textit{LTS measure} is defined as:
\begin{equation*}
LTS_{w,\sigma,\lambda,\zeta}(f,g)=\exp(-LTS_{w,\sigma}(f^\ast,g^\ast)/M-DP_{\lambda,\zeta}(g)).
\end{equation*}
\end{definition}
The scaling through the division by $M$ normalizes the LTS distance to the interval $[0,1]$. 
The transformation $[0,+\infty)\ni x\rightarrow\exp(-x)\in(0,1]$ maps the sum of the LTS distance and the duration penalty term to the interval $(0,1]$, while reversing the order as well: $g$ is closer to $f$ if the LTS measure is closer to~1.

\section{Application to activity recognition}

\subsection{Simulation study}
\label{sect:sim}

We consider a dataset created using a random procedure, which mimics the behavior of activity recognition classifiers with varying accuracy (depending on the parameters).
Let $\mathcal{S}=\{1,2,3\}$.
Consider a function $f$ representing a 60 second long state sequence:
\begin{equation*}
f=\mathbbm{1}_{[0,5)}+2\cdot\mathbbm{1}_{[5,15)}+3\cdot\mathbbm{1}_{[15,30)}+2\cdot\mathbbm{1}_{[30,40)}+3\cdot\mathbbm{1}_{[40,55)}+\mathbbm{1}_{[55,60]}.
\end{equation*}
$f$ will be referred to as the correct labels.
We introduce noise into $f$ in the following manner:
\begin{itemize}
\item two sequences of i.i.d. random variables are considered $\{Y_k\}$ and $\{Z_k\}$, with $Y_k\sim Exp(\mu_1)$ and $Z_k\sim Exp(\mu_2)$ for some parameters $\mu_1,\mu_2>0$,
\item $\{Y_k\}$ represents the time spent in the correct state, while $\{Z_k\}$ represents the time spent in the incorrect state,
\item we use the sequence $Y_1, Z_1, Y_2, Z_2, ...$ to generate noisy labels, where the sequence ends when the sum of all drawn numbers is exceeding 60 seconds,
\item for each variable $Z_i$ an incorrect state is chosen randomly out of the remaining two and $f$ is changed to that state on interval $[\sum\limits_{k=1}^{i-1}(Y_k+Z_k) +Y_i,\sum\limits_{k=1}^{i}(Y_k+Z_k))$,
\item $\mu_1$ and $\mu_2$ control the duration of the states.
\end{itemize}

As our performance measure we choose the LTS measure with parameters: $w = 0.6$, $\sigma = 0.35$, $\lambda = 0.0001$, $\zeta = 0.5$, $d=\rho$.
The post-processing is performed for the noisy labels with parameter $\gamma=0.5s$.
To demonstrate the utility of post-processing procedure, we draw the noisy functions 1000 times for a given set of parameters $(\mu_1,\mu_2)$ and compare the results between the accuracy of the noisy labels, the LTS measure of the noisy labels and the LTS measure of the post-processed labels.

\begin{figure}
\centering
\includegraphics[width=\linewidth]{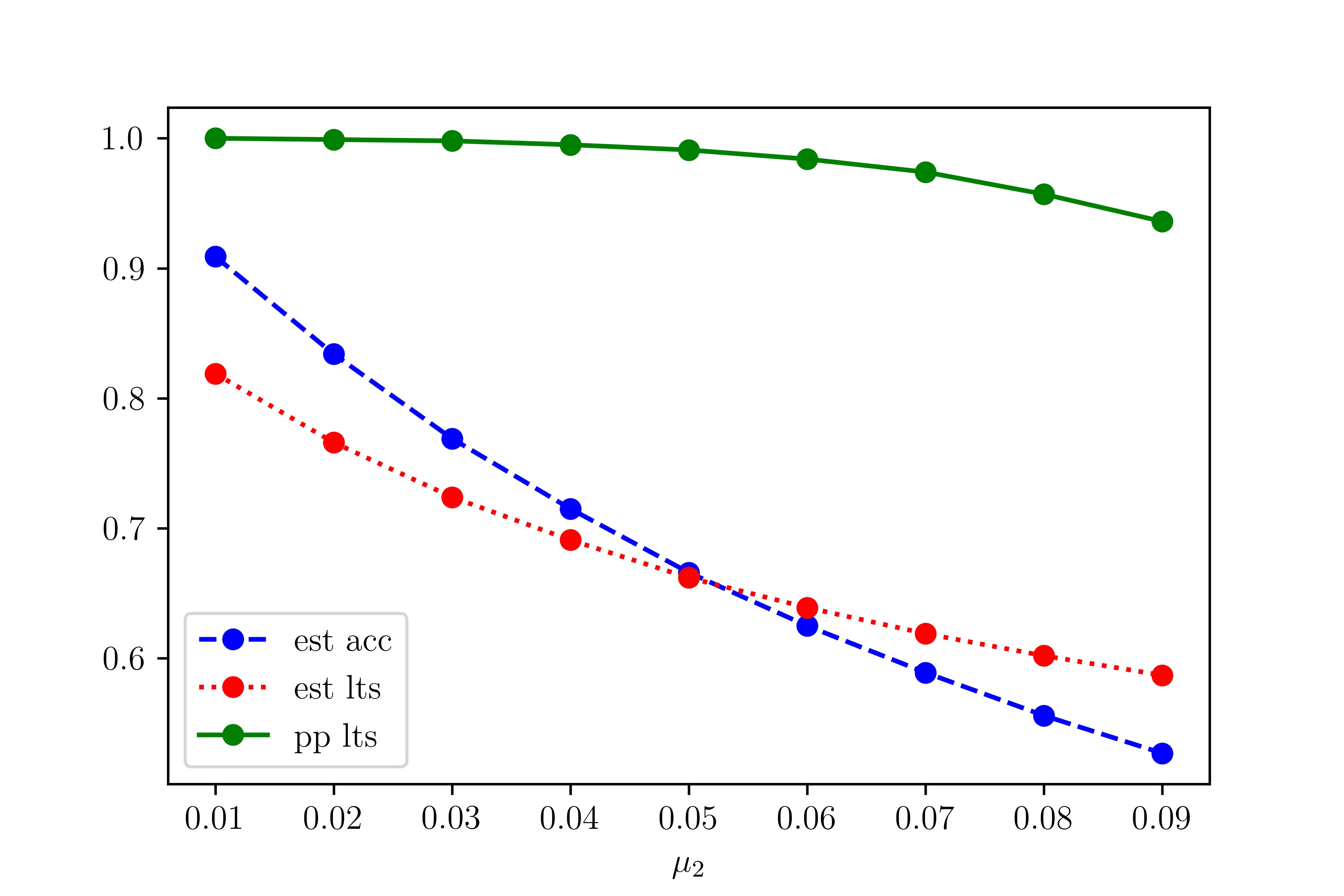}
\caption{The dashed line shows the mean accuracy of the noisy labels, the dotted line shows the LTS measure of the noisy labels and the solid line shows the LTS measure of the post-processed labels. All lines drawn for 9 different values of $\mu_2$.}
\label{fig:smallmu}
\end{figure}

In the first setting, we fix $\mu_1=0.1s$.
The procedure is repeated for $\mu_2\in\{0.01, 0.02, 0.03, 0.04, 0.05, 0.06, 0.07, 0.08, 0.09\}$.
Figure \ref{fig:smallmu} shows the mean accuracy of the noisy labels and the mean LTS measure of both the noisy labels and the post-processed labels.

\begin{figure}
\centering
\includegraphics[width=\linewidth]{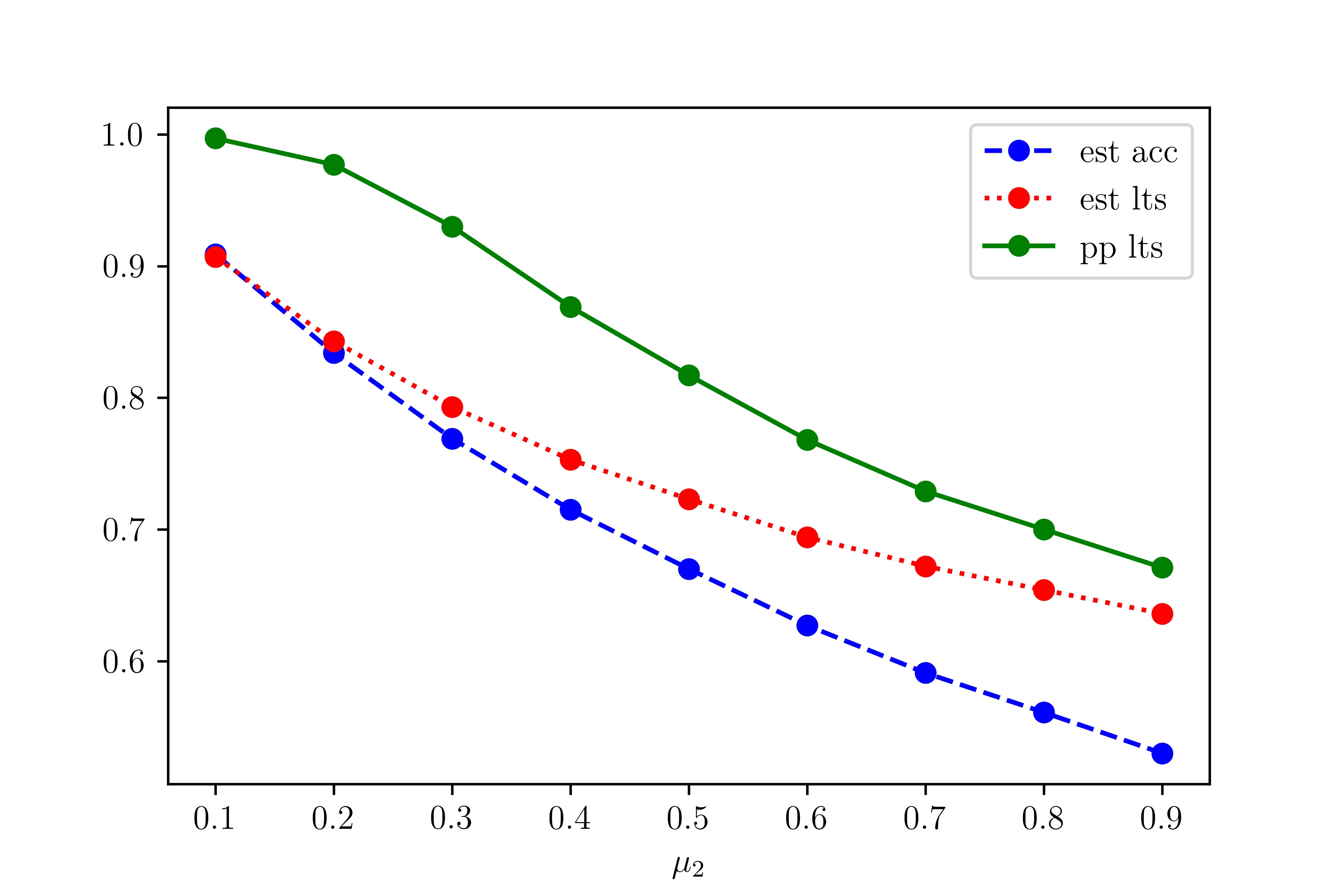}
\caption{The dashed line shows the mean accuracy of the noisy labels, the dotted line shows the LTS measure of the noisy labels and the solid line shows the LTS measure of the post-processed labels. All lines drawn for 9 different values of $\mu_2$.}
\label{fig:bigmu}
\end{figure}

In the second setting, we fix $\mu_1=1s$.
The procedure is repeated for $\mu_2\in\{0.1, 0.2, 0.3, 0.4, 0.5, 0.6, 0.7, 0.8, 0.9\}$.
Figure \ref{fig:bigmu} shows the mean accuracy of the noisy labels and the mean LTS measure of both the noisy labels and the post-processed labels.

Both experiments show the improvement in the LTS measure thanks to the use of post-processing.
Also, we conclude that the post-processing method behaves better when dealing with multiple shorter intervals rather than fewer longer ones.

\begin{figure}
\centering
\includegraphics[width=\linewidth]{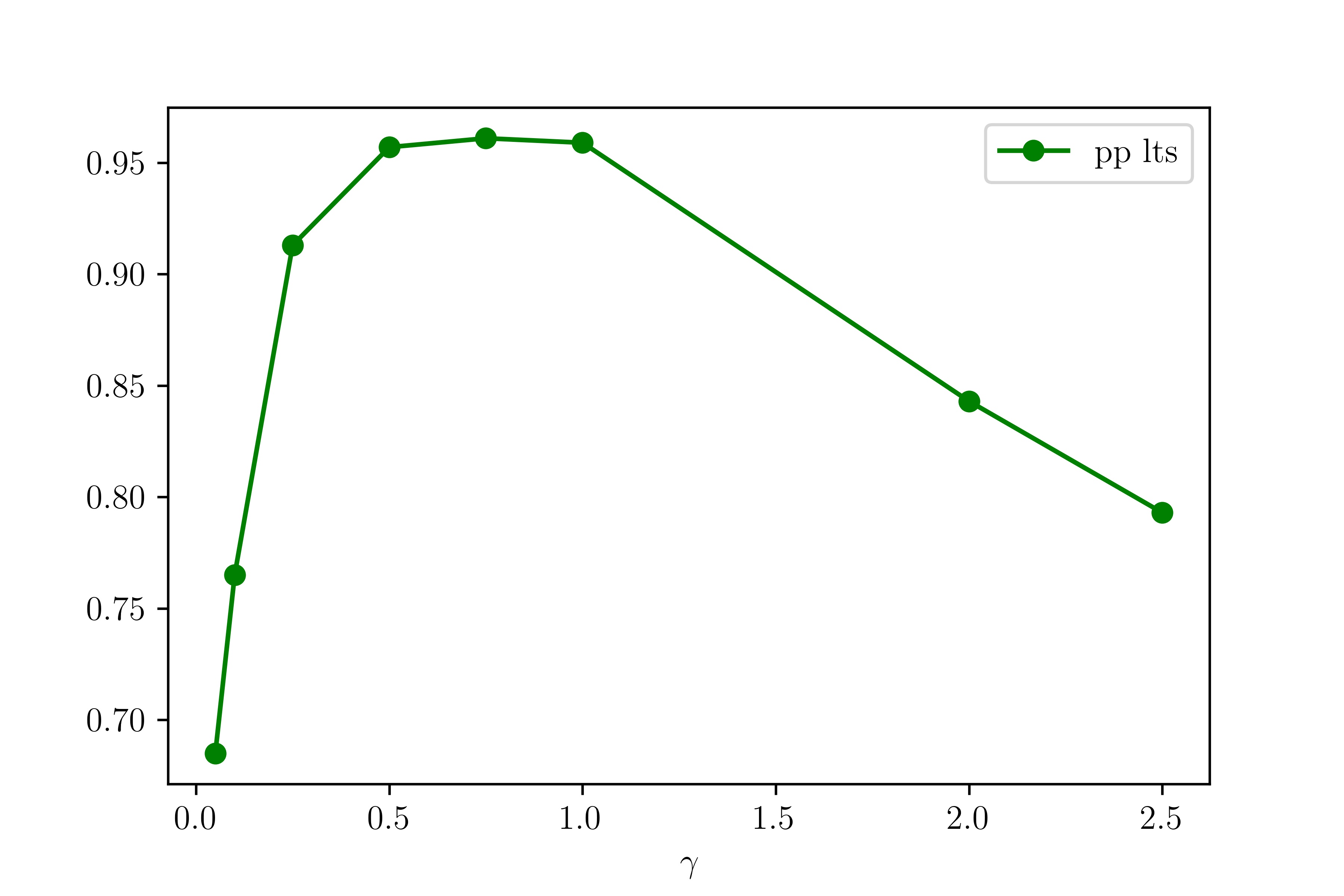}
\caption{The line shows the LTS measure of the post-processed labels drawn for 8 different values of $\gamma$. The mean accuracy of noisy labels was equal to 0.555 and the mean LTS measure of noisy labels was equal to 0.602.}
\label{fig:gamma}
\end{figure}

We also investigate the importance of $\gamma$ parameter on the results.
We fix $\mu_1=0.1$, $\mu_2=0.08$.
The procedure is repeated for $\gamma\in\{0.05, 0.1, 0.25, 0.5, 0.75, 1, 2, 2.5\}$.
Figure \ref{fig:gamma} shows the mean LTS measure of the post-processed labels.

We conclude that parameter $\gamma$ can influence the LTS measure of the post-processed functions $\hat{g}_i$. 
Careful consideration needs to be made as too low value of $\gamma$ will lead to accepting of unrealistically short events, while to high level of $\gamma$ will eliminate events in true labels. 
In our case the value of $\gamma$ between 0.5 and 1 is the most favourable. 
In practice the minimal length of the events in the true labels will inform on the choice of $\gamma$.

\begin{figure}
\centering
\includegraphics[width=\linewidth]{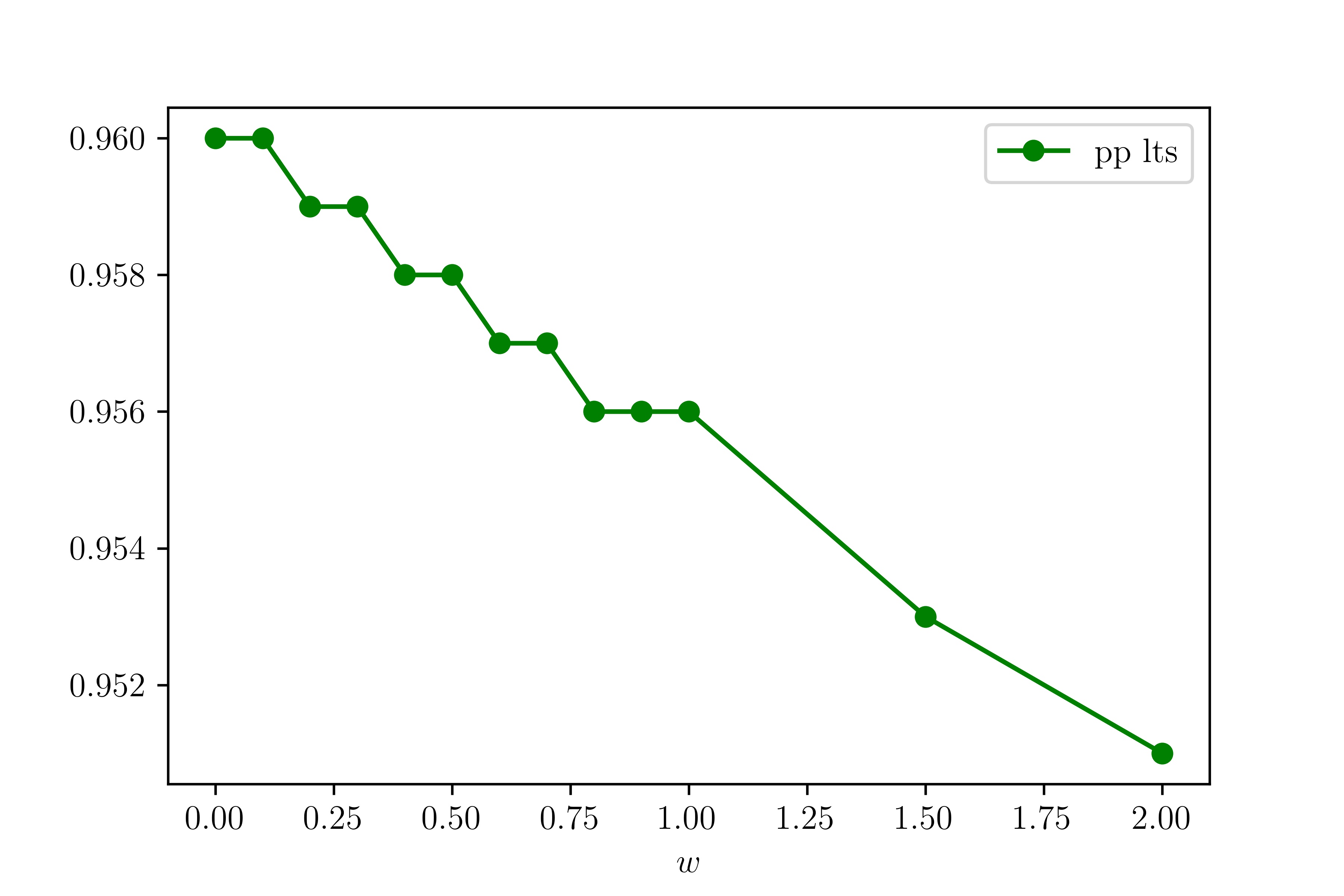}
\caption{The line shows the LTS measure of the post-processed labels drawn for 13 different values of $w$. The mean accuracy of noisy labels was equal to 0.555.}
\label{fig:w}
\end{figure}

We finish the simulation study with a look at the parameters of the LTS measure. 
We will investigate the weight $w$ first. 
Let all the other parameters of the LTS measure be set to $\sigma=0.35$, $\lambda=0.0001$, $\zeta=0.5$.
We fix $\mu_1=0.1, \mu_2=0.08, \gamma=0.5$.
The procedure is repeated for 13 different values of $w$.
Figure \ref{fig:w} shows the mean LTS measure of the post-processed labels.

We conclude that the choice of $w$ is of lesser importance as its effect on the LTS measure is minimal, since all the values on the $y$-axis of figure \ref{fig:w} are quite close together.

\begin{figure}
\centering
\includegraphics[width=\linewidth]{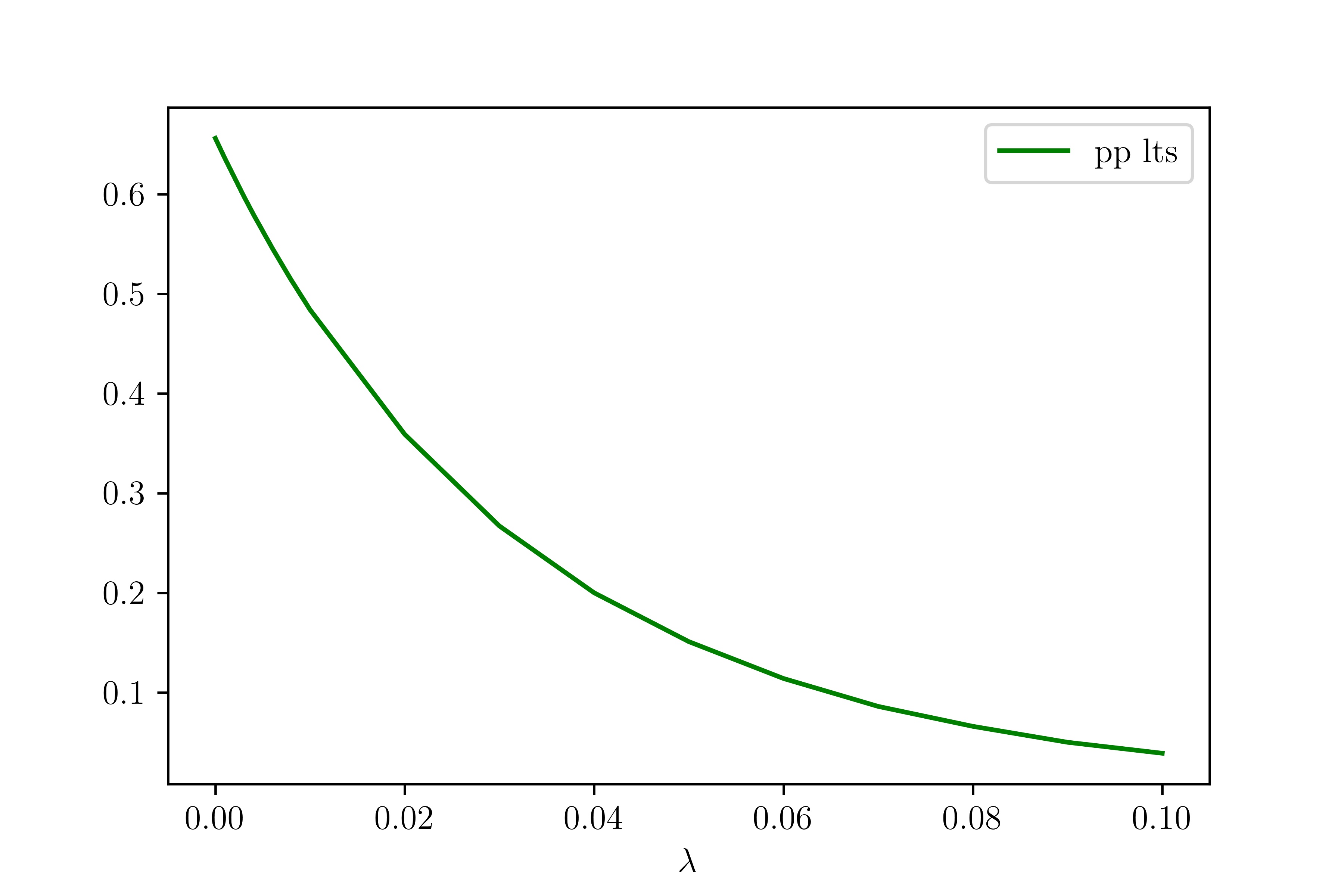}
\caption{The line shows the LTS measure of the post-processed labels drawn for $\lambda$. The mean accuracy of noisy labels was equal to 0.56.}
\label{fig:lambda}
\end{figure}

Parameters $\sigma$ and $\zeta$ will not be subjected to the same procedure, as they have a clear interpretation and can be chosen based on domain knowledge.
Hence, the only parameter left to investigate is $\lambda$.
As before, we fix $\mu_1=1, \mu_2=0.8, \gamma=0.5$.
We choose $w=0.6$.
The procedure is repeated for values of $\lambda$ between 0 and 0.1.
Figure \ref{fig:lambda} shows the mean LTS measure of the post-processed labels.
We can see that high values of $\lambda$ can influence the LTS measure significantly, hence choices lower than 0.01 are preferable.
We do not want the penalty term to be overshadowing the LTS distance.

\subsection{Application to football dataset}

We will now demonstrate the benefits of the post-processing by projection in a real-life setting, utilizing the LTS measure to compare different methods of classification.
\citet{wilmes20} give an extensive description of the football dataset of which we give a short summary below.

Eleven amateur football players participated in a coordinated experiment at a training facility of the Royal Dutch Football Association of The Netherlands.
Five Inertial Measurement Units (IMUs) were attached to 5 different body parts: left shank (LS), right shank (RS), left thigh (LT), right thigh (RT) and pelvis~(P).
Each IMU sensor contains a 3-axis accelerometer (Acc) and a 3-axis gyroscope (Gyro).
Athletes were asked to perform exercises on command, e.g. `jog for~10 meters' or `long pass'. 
For each athlete and exercise this resulted in a 30-dimensional time series (5 body parts times 6 features per IMU) of length varying from 4 to 14 seconds.
Each athlete performed 70-100 exercises which amounts to nearly~900 time series (each with a sampling frequency of 500 Hz).
Time series are labelled with the command given to an athlete, but there are still other activities performed in each of the time series, for example standing still.
This causes a problem; ignoring standing periods and treating them as part of the main signal pollutes the data and lowers the quality of the classification.
To show the advantages of post-processing by projection, we select only two states: `standing' and `other activity' encoded as $0$ and $1$, respectively.
15 time series (representative of all possible actions performed by athletes) were manually labelled time point by time point in order to be able to train classifiers, and these will form our sample.

In pre-processing we are using the sliding window technique on the sensors \citep{dietterich02}. 
This method transforms the original raw data using windows of fixed length $d$ and a statistic of choice $T$: given a time point $t$, its neighbourhood of size $d$ is fed to the statistic $T$ for each variable separately.
Performing the procedure for each time point results in a time series of the same dimension as the original one, but every observation is equipped with some knowledge about the past and the future through the statistic $T$ and through forming the neighbourhoods of size $d$.
Regarding the choice of the statistic $T$ one needs to be careful, since the sensors are highly correlated with each other.
The information about standing contained in one variable is comparable to the one in another, namely the variance of the signal is low when the person is standing (differences can occur when considering different legs; a low variance on one leg might be misleading since the other leg might already be transitioning into another position). 

10-fold cross-validation will be performed in order to select the best performing classification method out of the 7 standard machine learning methods, which will be mentioned later in the paper.
A typical approach to $k$-fold cross-validation with a training sample of size $k-1$ cannot be applied here, since a single time series is not a representative sample of different types of events.
15 time series will be used.
In each iteration 10 time series will be randomly chosen for training and 5 for testing.
The results are going to be shown for post-processed classifiers, unless specified otherwise.
Before cross-validation can be performed, we need to fix the parameters of the performance measure we introduced in section 2.
The parameters of the LTS measure are chosen as follows:
\begin{itemize}
    \item We have limited information regarding how uncertain locations of state transitions are, but based on the small experiment described in section~\ref{firstrealsect} we select $\sigma=0.35$ (the largest deviation between different true labels).
    \item The parameter $w$ is chosen as $0.6$, but as shown in section \ref{sect:sim} its choice is not that important.
    \item The lower bound $\gamma$ on the duration of activities is selected as the length of the shortest activity in the learning dataset, which is equal to $0.8$s in our case.
    \item A penalty $\lambda$ represents the cost of additional or missing jumps in a state sequence compared to the true labels.
    We decide for the penalty~$\lambda=0.01$ in order not to overshadow the LTS distance with too much importance placed on the penalty term (more details on that in section \ref{sect:sim}, specifically figure \ref{fig:lambda}).
\end{itemize}

Before assessing classifiers on the training set, one needs to consider an appropriate feature set.
Our variables are highly dependent on one another, so we start with feature selection.
We perform feature ranking using the Relieff algorithm and select the 6 most relevant features based on the Relieff weights (more details on the method in \citet{kononenko97}). 
Then we test all possible combinations of these features, which is now computationally feasible, in order to find the best set for each of the classifiers. 
The features selected by the Relieff algorithm are RTGyroX, RTGyroY, RTAccX, RTAccZ, LTAccY, PAccY, where the naming convention is as follows: RTGyroX refers to the $x$-axis of the gyroscope located on the right thigh.

\begin{table}
\begin{center}
\begin{tabular}{lccc}
\toprule
Classifier & OG Test & PP Test  \\
\midrule
MLP & 0.916+/-0.031 & 0.972+/-0.008 \\
LR & 0.898+/-0.034 & 0.968+/-0.015 \\
kNN & 0.59+/-0.05 & 0.967+/-0.020 \\
RF & 0.83+/-0.07 & 0.966+/-0.017 \\
SVC & 0.894+/-0.034 & 0.966+/-0.017	\\
DT & 0.83+/-0.07 & 0.965+/-0.008 \\
NB & 0.88+/-0.04 & 0.944+/-0.023 \\
\bottomrule
\end{tabular}
\end{center}
\caption{Average of the 10-fold cross-validation scores for all classifiers using the best sensor set for each of them.
The pre-processing consisted of the sliding window technique in combination with summarizing by the standard deviation.
The OG Test averages the LTS measure on the test set for the original classifier, while the PP Test is the same value for the post-processed classifier.}
\label{CrossvalALL}
\end{table}

Proceeding with the cross-validation we select the following classifiers (with their abbreviations) to be assessed: DT - Decision Tree, kNN - k-Nearest Neighbors, LR - Logistic Regression, MLP - Multi-layer Perceptron, NB - Naive Bayes, RF - Random Forest, SVM - Support Vector Machine. 
The results of 10-fold cross-validation are shown in table \ref{CrossvalALL}.
It is striking that the test scores of the post-processed classifiers are at most 0.028 apart.
This is due to post-processing by projection.
The correction it provides brings all classifiers closer together.
This astonishing result can be extended even further.
The test score of a decision tree ranges from 59\% to 86\% for different sensor sets before post-processing, while using the post-processing results in a range of test scores from 93\% to~96.5\% and this is not specific to decision trees only.

The example shows that the post-processing is crucial. 
Firstly, it increases the accuracy of a given estimator on a given feature set by 35\%.
Secondly, it diminishes the impact of feature selection as the difference in accuracy between different feature subsets decreases substantially.
Feature selection is of course still important as it decreases computational complexity of the problem and allows to get rid of redundancy in the feature set.
However, with methods that only rank features such as Relieff the choice of the threshold we choose to classify a feature as significant or not is less important.
Finally and most importantly, the post-processing by projection allows to select a method according to criteria other than the performance, namely the computational speed.

\section{Conclusion}

In this paper we have introduced a post-processing scheme that allows to improve estimates.
It finds estimated activities that are too short and eliminates them in an optimal way by finding the shortest path in a directed acyclic graph.

A simulation study is conducted to assess the benefits brought by the post-processing method.
Generated noisy labels are improved with the use of the post-processing.
The positive effects on the LTS measure are more significant when the noisy sequence contains more shorter intervals of misclassification.

Real-life football sensor data were used to assess the adequacy of the post-processing scheme in the more realistic setting.
It significantly improved the performance of the classifiers.
At the same time, post-processed classifiers are closer to each other in performance than the original ones.
This allows placing more importance on other criteria, such as the computational speed of a method.
It should be noted that post-processing cannot correct for uncertainty in the classification result of the estimators.
It can be seen in figures \ref{fig:smallmu} and \ref{fig:bigmu} that the worse the original estimate the worse the post-processed one (at least as a rule of thumb as there can be cases when it is reversed).
However, most importantly, the results of the application to the football dataset are promising.
The post-processing by projection was able to improve the estimators of accuracy ranging from 59\% to 86\% up to a score of 93\% to 96.5\%.
It is notable that the lowest score that the post-processed estimates have achieved for a given classification method is still higher than the highest score of the original estimates.

Our second contribution are novel measures of classifier performance in the task of activity recognition using wearable sensors.
They address the issue of timing offsets as well as unrealistic classifications, while retaining a typical scalar output of a performance measure allowing for easy comparisons between classifiers.

\section*{Acknowledgments}

We thank Erik Wilmes for providing football data of high quality and the stick-model animation tool. 
It was the basis for the analysis of our methods in section~4. 
We also thank Bart van Ginkel for the idea of how to generalize the performance measure from the binary to the multiclass case.

\section*{Declarations}

\subsection*{Author contributions}
All authors contributed to the study conception and design. Data analysis were performed by 
the first author. 
The first draft of the manuscript was written by 
the first author and all authors commented on previous versions of the manuscript. All authors read and approved the final manuscript.

\subsection*{Availability of data and material}
The dataset analysed during the current study are available under the link \url{https://zenodo.org/record/3732988\#.YMcXOqgzZEZ}.

\subsection*{Code availability}
Custom code for post-processing and performance measures can be found at: \url{https://github.com/mgciszewski/improving_state_estimation_2022}.

\subsection*{Declaration of interests}
The authors declare that they have no known competing financial interests or personal relationships that could have appeared to influence the work reported in this paper.

\subsection*{Funding}
\begin{minipage}[t]{.7\columnwidth}
This work is part of the research programme CAS with project number P16-28 project 2, which is (partly) financed by the Dutch Research Council (NWO).
\end{minipage}
\hspace{0.02\linewidth}
\begin{minipage}[t]{.2\linewidth}
  \raisebox{-\height+0.7\baselineskip}{\includegraphics[width=0.9\linewidth]{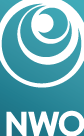}}
\end{minipage}

\begin{appendix}

\section{Proofs}

\paragraph{Proof of lemma \ref{jumpsoutside}}
Let $\hat{f}$ be a solution of the problem \ref{eq:projectiondist} for a given function $f$.
Assume that $\hat{f}$ contains a jump $t$ outside of the set $J(f)$.
Denote the jump or one of the jumps closest to $t$ in the original function $f$ by~$t_k$. 
Without loss of generality we assume $t_k$ is located left of $t$ ($t_k$ exists otherwise~$f$ is constant and $\hat{f}=f$).
Let $t_a$ and $t_b$ denote the jump preceding and resp. following $t$ in the projection $\hat{f}$.
Let $t_{k+1}$ denote the jump following $t_k$ in the original function $f$.
Let $s_1$ be the state in which the original function stays in the interval $[t_k,t_{k+1})$ and let $s_2$ be the state from which the projection $\hat{f}$ jumps at $t$ and let $s_3$ be the state to which the projection $\hat{f}$ jumps at $t$.

\begin{figure}
\centering
\includegraphics[width=0.75\linewidth]{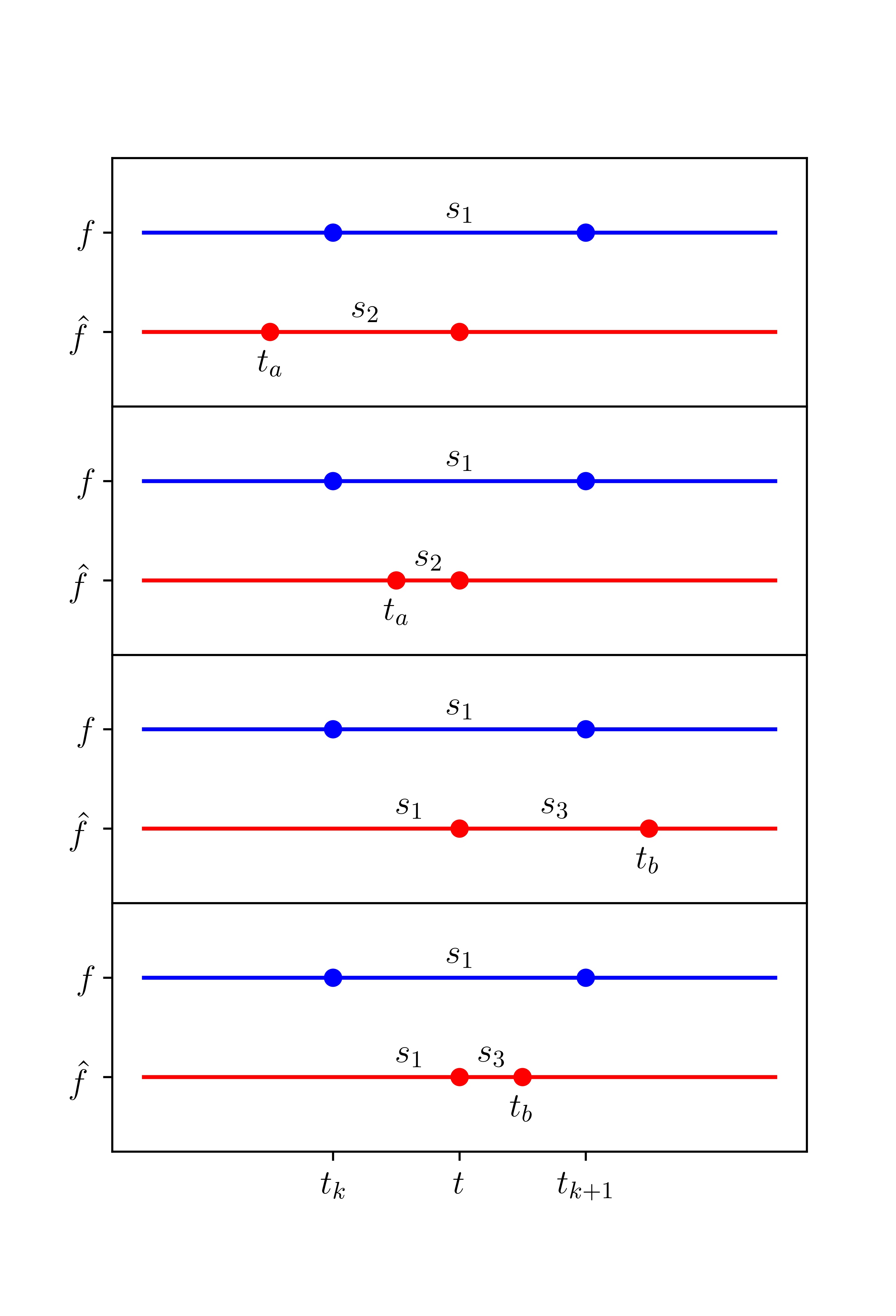}
\caption{Illustration supporting the proof of lemma \ref{jumpsoutside}. Plots correspond (from top to bottom) to cases 1a, 1b, 2a, 2b respectively}
\label{fig:proofgraphs}
\end{figure}

We will consider multiple cases and in each of them we will present a modification to $\hat{f}$ that either shows that $\hat{f}$ cannot be a projection or that there exists a function which is not worse than $\hat{f}$ and does not contain a jump at $t$. The configurations of the cases are depicted in Fig. \ref{fig:proofgraphs}.
\begin{enumerate}
\item $s_1\neq s_2$
\begin{enumerate}
\item $t_a<t_k$. Moving the jump $t$ to $t_k$ does not increase the error (and potentially lowers it, if $s_3=s_1$).
\item $t_a\geq t_k$. We move the jump $t_a$ to $t$, which results in lowering the error by penalty term $\gamma$. Then we go back to the beginning of the proof with redefined state $s_2$ and jump $t_a$.
\end{enumerate}
\item $s_1=s_2$
\begin{enumerate}
\item $t_b\geq t_{k+1}$. Moving the jump $t$ to $t_{k+1}$ lowers the error by $t_{k+1}-t$ since $s_3\neq s_1$.
\item $t_b<t_{k+1}$. We move the jump $t$ to $t_b$, which results in lowering the error by penalty term $\gamma$ and $t_b-t$, since $s_1=s_2$. We go back to the case $s_1=s_2$ with $t_b$ redefined as the jump following the newly defined $t$. Eventually the jump $t$ can be replaced by $t_{k+1}$ (when case 2(a) is reached).
\end{enumerate}
\end{enumerate}
The loop occurring in case 1(b) is not problematic, since with each iteration the number of jumps of the solution is reduced, eventually case 1(a) or 2 is reached.\qed

\paragraph{Proof of lemma \ref{gammainterpret}}
Let $\hat{f}$ be a solution of the problem \ref{eq:projectiondist} for a given function $f$.
Assume that for certain $\tilde{\gamma}<\gamma$, $\hat{f}\in\mathcal{G}_{\tilde{\gamma}}$ and $\hat{f}\not\in\mathcal{G}_{\gamma}$.
Hence there exist two jumps $t_k$ and $t_l$ of $f$ and $\hat{f}$ (which follows from lemma \ref{jumpsoutside}), such that $\tilde{\gamma}<t_l-t_k<\gamma$.
Since the state lasts less than $\gamma$, it can be removed (in the sense that one of the jumps is removed and either the previous state or following state is longer by $t_l-t_k$) with a gain in error of less than $\gamma$ and decrease in error of exactly $\gamma$, which means we found a function with lower error than $\hat{f}$. 
This contradiction ends the proof.\qed

\paragraph{Proof of lemma \ref{longstates}}
Let $f$ be a function with two neighboring jumps $t_1,t_2$ and the state $s_1$ between them.
Assume $t_2-t_1\geq2\gamma$.
Since the interval is longer than or equal to $2\gamma$ it satisfies the condition of the class~$\mathcal{G}_\gamma$.
Let us assume that the projection $\hat{f}$ of~$f$ contains two neighbouring jumps $t_a$ and $t_b$ such that $t_a\leq t_1<t_2\leq t_b$ and the state in the interval $[t_a,t_b)$ is $s_2\neq s_1$.
We introduce notation $\alpha:=t_1-t_a$ and $\beta:=t_b-t_2$.
If $\alpha,\beta\geq\gamma$, then introducing the jumps at $t_1$ and $t_2$ with the state $s_1$ between them is possible, because the condition of the class~$\mathcal{G}_\gamma$ is satisfied. Moreover, the error is decreased if $t_2-t_1>2\gamma$ and is not increased if $t_2-t_1=2\gamma$.
If $\alpha\geq\gamma$ and $\beta<\gamma$, then introducing a jump at $t_1$ such that the state following it is $s_1$ is possible. Moreover, the error is decreased. Analogously when $\alpha<\gamma$ and $\beta\geq\gamma$.
If $\alpha,\beta<\gamma$, then changing state $s_2$ to $s_1$ reduces the error.

In all cases, we have shown that there exists a projection that does not change the state longer than $2\gamma$.\qed

\paragraph{Proof of remark \ref{simplefact}}

Let $\hat{f}$ be a projection of $f$ onto $\mathcal{G}_\gamma$.
Let $t_1$ and $t_2$ be the first two jumps in the original function~$f$.
Let $s_1$ and $s_2$ be the first two states in the original function~$f$.
Assume that $t_2-t_1<2\gamma$.
If $\hat{f}$ had a jump at $t_2$ from the state~$s_1$, then a function $g$ equal to $\hat{f}$ outside of interval $[t_1,t_2)$, but such that the jump from state $s_1$ is moved to the location of the jump $t_1$ has the same or lower error than $\hat{f}$.
If $\hat{f}$ had a jump at $t_2$ from a state $s_i\neq s_1$, then the error is infinite (since the value of $\hat{f}$ differs from $f$ on the interval $(-\infty, t_1)$) and $\hat{f}$ cannot be a projection.

The argument is analogous for the penultimate jump.\qed

\paragraph{Proof of theorem \ref{equivtheorem}}

We use lemma \ref{jumpsoutside} to prove that a projection of a function from $\mathcal{T}$ onto $\mathcal{G}_\gamma$ can only have jumps at the same positions as the jumps in the original function.
This leads to the fact that finding the shortest path in the graph is equivalent to finding $\hat{f}$.\qed

\paragraph{Proof of lemma \ref{binarydirections}}

Let $\hat{f}$ be a projection of $f$ onto $\mathcal{G}_\gamma$.
Let $t_k$ and $t_{k+1}$ be two consecutive jumps of $f$.
Assume that $\hat{f}$ contains a jump $t_k$, but in opposite direction than in $f$.
From lemma \ref{jumpsoutside} we know that the next jump of $\hat{f}$ can occur at the earliest at $t_{k+1}$.
This means that in the interval $[t_k,t_{k+1})$ the projection $\hat{f}$ is equal to $1-f$.
In this case, moving the jump at $t_k$ to $t_{k+1}$ (or in the case of $t_{k+1}\in J(\hat{f})$ removing both jumps) reduces the error by $t_{k+1}-t_k$.
Hence, we conclude, a jump from $f$ can only be present in its projection if it is in the same direction as in $f$.\qed

\paragraph{Proof of lemma \ref{binarygammainterpret}}

The proof of this lemma is analogous to the proof of lemma \ref{gammainterpret}. The possibility of strengthening the previous result comes from the fact that we can remove two jumps at once, in effect reducing the error by $2\gamma$.

\paragraph{Proof of lemma \ref{binarylongstates}}

The proof of this lemma is analogous to the proof of lemma \ref{longstates}

\paragraph{Proof of theorem \ref{binaryequivtheorem}}

We use lemmas \ref{jumpsoutside} and \ref{binarydirections} to prove that a projection of a function from $\mathcal{T}$ onto~$\mathcal{G}_\gamma$ can only have jumps at the same positions and in the same directions as the jumps in the original function.
This leads to the fact that finding the shortest path in the graph is equivalent to finding $\hat{f}$.\qed

\paragraph{GTS distance with $w>0$ and $\sigma=\infty$ is an extended metric}
We will show that:
\begin{equation*}
GTS_w(f,g)=\inf\limits_{\epsilon\in\mathbb{R}}\{\textrm{dist}(f\circ\tau_\epsilon, g)+w\vert\epsilon\vert\}
\end{equation*}
is an extended metric on $\mathcal{T}$.
\begin{enumerate}
\setcounter{enumi}{-1}
\item Since for any $\epsilon$, $\textrm{dist}(f\circ\tau_\epsilon, g)\geq 0$ and $w\vert\epsilon\vert\geq 0$ we conclude that the $GTS_w$ is non-negative.
\item It is obvious to see that $GTS_w(f,f)=0$ for any $f\in \mathcal{T}$.
Now let us assume that for some $f,g\in \mathcal{T}$ we have $GTS_w(f,g)=0$.
This implies that
\begin{equation*}
\exists_{(\epsilon_n)}\;\;\textrm{dist}(f\circ\tau_{\epsilon_n}, g)+w\vert\epsilon_n\vert\xrightarrow{n\rightarrow\infty}0.
\end{equation*}
Since $\textrm{dist}(f\circ\tau_{\epsilon_n}, g)+w\vert\epsilon_n\vert$ is an upper bound of $\textrm{dist}(f\circ\tau_{\epsilon_n}, g)$ and $w\vert\epsilon_n\vert$, we have
\begin{align*}
\vert\epsilon_n\vert&\xrightarrow{n\rightarrow\infty}0,\\
\int\limits_\mathbb{R}d(f\circ\tau_{\epsilon_n}(t), g(t))d\lambda(t)&\xrightarrow{n\rightarrow\infty}0.
\end{align*}
From Fatou's lemma we have
\begin{equation*}
\int\limits_\mathbb{R}\liminf\limits_{n\rightarrow\infty}d(f(t-\epsilon_n),g(t))d\lambda(t)=0,
\end{equation*}
where $\lambda$ is the Lebesgue measure on $\mathbb{R}$.
Because $f$ and $g$ are càdlàg, this implies that for almost all $t$ we have $f(t-)=g(t)$ or $f(t)=g(t)$ and so we conclude that $f=g$.
\item Let $f,g\in \mathcal{T}$, we have
\begin{align*}
GTS_w(f,g)=&\inf\limits_\epsilon\{\textrm{dist}(f\circ\tau_\epsilon,g)+w\vert\epsilon\vert\}=\inf\limits_\epsilon\{\textrm{dist}(g\circ\tau_{-\epsilon},f)+w\vert-\epsilon\vert\}\\
=&\inf\limits_{-\epsilon}\{\textrm{dist}(g\circ\tau_{\epsilon},f)+w\vert\epsilon\vert\}=\inf\limits_{\epsilon}\{\textrm{dist}(g\circ\tau_{\epsilon},f)+w\vert\epsilon\vert\}\\
=&GTS_w(g,f),
\end{align*}
hence we conclude that $GTS_w$ is symmetric.
\item Letting $f,g,h\in \mathcal{T}$, we have
\begin{align*}
GTS_w(f,g)=&\inf\limits_\epsilon\{\textrm{dist}(f\circ\tau_\epsilon,g)+w\vert\epsilon\vert\}\\
=&\inf\limits_{\epsilon_1,\epsilon_2}\{\textrm{dist}(f\circ\tau_{\epsilon_1}\circ\tau_{\epsilon_2},g)+w\vert\epsilon_1+\epsilon_2\vert\}\\
\leq&\inf\limits_{\epsilon_1,\epsilon_2}\{\textrm{dist}(f\circ\tau_{\epsilon_1}\circ\tau_{\epsilon_2},h\circ\tau_{\epsilon_2})+\textrm{dist}(h\circ\tau_{\epsilon_2},g)+\\
&+w\vert\epsilon_1\vert+w\vert\epsilon_2\vert\}\\
=&\inf\limits_{\epsilon_1,\epsilon_2}\{\textrm{dist}(f\circ\tau_{\epsilon_1},h)+w\vert\epsilon_1\vert+\textrm{dist}(h\circ\tau_{\epsilon_2},g)+w\vert\epsilon_2\vert\}\\
=&\inf\limits_{\epsilon_1}\{\textrm{dist}(f\circ\tau_{\epsilon_1},h)+w\vert\epsilon_1\vert\}+\inf\limits_{\epsilon_2}\{\textrm{dist}(h\circ\tau_{\epsilon_2},g)+w\vert\epsilon_2\vert\}\\
=&GTS_w(f,h)+GTS_w(h,g),
\end{align*}
which shows that $GTS_w$ satisfies the triangle inequality and that concludes the proof.\qed
\end{enumerate}

\paragraph{The LTS distance with $w>0$ is a semimetric}
Let $w>0,\sigma>0$ and a metric $d$ on $\mathcal{S}$ be fixed.
We observe that $LTS_{w,\sigma}$ is nonnegative.
Symmetry of $LTS_{w,\sigma}$ follows directly from the definition.
It only remains to show that $LTS_{w,\sigma}(f,g)=0$ if and only if $f=g$ for $f,g\in\mathcal{T}$.

We have
\begin{equation*}
LTS_{w,\sigma}(f,f)=0,
\end{equation*}
because there is only one segment (as defined in \ref{segments}).
Assume now that $LTS_{w,\sigma}(f,g)=0$ and $f\neq g$.
In that case, there exists more than one segment.
\begin{align*}
LTS_{w,\sigma}(f,g)&=\sum\limits_{i=1}^{l-1}\delta_i(a_{i+1}-a_i)d(f(a_i),g(a_i))=0\\
&\Rightarrow\forall_{i=1,2,3,...,l-1}\quad f(a_i)=g(a_i),
\end{align*}
which implies that $f=g$, which contradicts the assumption.
We conclude that $LTS_{w,\sigma}(f,g)=0$ iff $f=g$, which completes the proof.\qed

\end{appendix}

\bibliography{Manuscript}

\end{document}